\documentclass[runningheads]{llncs}
\usepackage[T1]{fontenc}

\usepackage{multirow}
\usepackage{graphicx}
\usepackage{booktabs}
\usepackage{setspace}
\usepackage{longtable}
\usepackage{enumitem}

\title{Prediction of 30-day hospital readmission with clinical
notes and EHR information}

\author{Tiago Almeida\inst{1} \and Plinio Moreno \inst{1}\orcidID{0000-0002-0496-2050} \and 
Catarina  Barata\inst{1}\orcidID{0000-0002-2852-7723}}
\authorrunning{T. Almeida et al.}
\institute{Institute for Systems and Robotics, LARSyS, Instituto Superior Técnico, Portugal }

\begin{document}

\maketitle

\begin{abstract}
High hospital readmission rates are associated with significant costs and health risks for patients. Therefore, it is critical to develop predictive models that can support clinicians to determine wether or not a patient will return to the hospital in a relatively short period of time (e.g, 30-days). Nowadays, it is possible to collect both structured (electronic health records - EHR) and unstructured information (clinical notes) about a patient hospital event, all potentially containing relevant information for a predictive model. However, their integration is challenging. In this work we explore the combination of clinical notes and EHRs to predict 30-day hospital readmissions. We address the representation of the various types of information available in the EHR data, as well as exploring LLMs to characterize the clinical notes. We collect both information sources as the nodes of a graph neural network (GNN). Our model achieves an AUROC of 0.72 and a balanced accuracy of 66.7\%, highlighting the importance of combining the multimodal information. 


\end{abstract}

\section{Introduction}
\label{sec:intro}

Hospital readmission rates are regarded an indicator of hospital quality of care \cite{Ashton1997}, and are frequently used as a metric to evaluate the performance of healthcare systems \cite{Jencks2009}. While readmissions are particularly relevant to insurance companies, as their core business is supporting patient's health care bills, patients' health is the real concern. Often, readmissions are associated with the worsening of the initial conditions or the development of new complications, which may lead to longer hospital stays, higher treatment expenses, and even mortality \cite{Glans2020}.

Machine learning (ML) models have been gaining interest as a suppport tool that can help healthcare professionals in patient monitoring and decision making. These models can help on identifying patients at high risk of readmission, allowing professionals to intervene and act accordingly to help the patient, by providing the necessary care and monitoring. At first, prediction models were mainly based on structured data (Electronic Health Records - EHR), which include information as demographics, comorbidities, medication, diagnostic codes, and vital signs. However, this data offers a limited perspective of the patient's journey, as it lacks detailed information about the care provided during the hospital stay. Clinical notes are essential to understand these details, as they are free text narratives written by healthcare professionals that describe changes in the health condition, the care provided, and the patient's response to treatment. Yet, clinical notes are unstructured data, which makes it difficult to extract useful information from them. They comprise highly specialized terminologies, abbreviations, colloquialisms, and jargon, but also inconsistencies, such as misspellings, nomenclature conventions, transcription errors, and unusual grammatical structure. Therefore, the seamless integration of EHRs and clinical notes into a single model remains a challenge. 

In this work, we propose a novel strategy to combine a large body of information contained in the EHRs with clinical notes, towards the prediction of 30-day hospital readmissions. To integrate the two modalities, we explore graph neural networks (GNNS), treating each admission as a node in a graph and linking togethet admissions with similar characteristics. This allows us to handle a large cohort of patients (over 100,000), by capturig similarities between admissions.

\section{Related Work}
\subsection{Background}
Conventional machine learning approaches have been applied to hospital readmission prediction \cite{Huang2021,Ben-Assuli2018}. These models primarily use structured EHR data, which includes demographics, comorbidities, medication, diagnostic codes, and vital signs. However, these models often struggle with capturing complex relationships and temporal dependencies in the data. Commonly used models include:
Logistic Regression (LR), Random Forests (RF), Support Vector Machines (SVM), Boosted Decision Trees (BDT), LASSO, Ridge Regression, and Elastic-Net. These models have shown promising results in predicting hospital readmissions, with AUROC scores ranging from 0.6 to 0.8 \cite{Tong2016}. However, these models are relying just on static information from the EHR data, that can be limiting in capturing the temporal dependencies in the data and very high-dimensional data. This has led to the development of more advanced models, such as NNs and Deep Learning (DL) models, that can take advantage of the sequential nature of the data and learn complex patterns and relationships in the data.

\subsection{Neural Network architectures and Deep Learning Models}
Deep learning (DL) models have shown promising results in predicting hospital readmission by leveraging various types of medical data, including EHR information, clinical images and notes. Models such as Multi-Layer Perceptron (MLP), Convolutional Neural Networks (CNN) and Transformers have been applied to readmission prediction. In the case of MLP, Hai \textit{et al.} \cite{Hai2023} focuses on readmission of diabetes patients, attaining an AUROC of 0.69, similar to the LR model. In the
same study, LSTM had the best performance with an AUROC score of 0.79,
followed by AdaBoost, a boost ensemble method, with a score of 0.72. Zebin \textit{et al.} \cite{Zebin2019} proposed an heterogeneous LSTM+CNN model, where the LSTM processes the EHR information and then the Convolution Neural Network (CNN) computes the feature maps that are used in the output decision layer. Unlike RNNs, LSTMs, and CNNs, Transformers \cite{Devlin2019} are able to weight the importance of different parts in the input, making them specially useful for clinical tasks that require understanding the context of the prediction. \cite{Huang2020} relies on a Bidirectional encoder representations from transformers (BERT) model, which is trained with Clinical Notes. This model is fine-tuned to predict hospital readmissions during different admission phases, instead of only at discharge, which allows to intervene earlier. Alsentzer \textit{et al.} \cite{Alsentzer2019} used a similar approach, by pretraining BERT and BioBERT \cite{Lee2019} models on Clinical Notes at discharge and all times. These models were evaluated against the base models, and concluded that their models achieved greater cohesion around medical or clinic operations relevant terms than the base models. These studies represent a big part on the recent advances in hospital readmission prediction, being the current state-of-the-art models for text-based clinical tasks.

\subsection{Graph Neural Networks}
\label{section:graph-neural-networks}

Graph Neural Networks (GNNs) are able to model complex relationships and dependencies, such as molecular interactions, patient networks, or clinical pathways. For hospital readmission prediction, GNNs have been used to model the relationships between patients, diagnoses, treatments, and admissions, allowing for predictions that take into account the complex dependencies between these entities.
\cite{Golmaei2021} proposed a model that combined clinical notes information and a patient network topological structure to predict hospital readmissions. Their framework consists of two main components: DeepNote, which extracts deep representations of clinical notes using ClinicalBERT \cite{Huang2020} with a feature aggregation unit; and a Graph Convolutional Network (GCN) that builds a patient network and trains it for hospital readmission predictions, with each node representing a patient's admission using the deep note representation and each edge representing the similarity between two admissions. The model based on discharge notes achieved an AUROC score of 0.858. They also reported that their network contained 6,162 admission nodes and 3,667,733 edges with the cosine similarity threshold adjusted to 0.99. DeepNote-GNN was able to model the complex relationships between patients and admissions making it a powerful tool for hospital readmission prediction. However, the model has limitations, such as only using clinical notes and not considering the temporality of hospital admission data.

Recently, Tang \textit{et al.} \cite{Tang2023} proposed a Multimodal Spatiotemporal Graph Neural Network (MM-STGNN) for prediction of hospital readmission that uses chest radiographs and EHR data, that achieved an AUROC score of 0.79. Their model consists of two subnetworks that process the chest radiographs and EHR data separately, and a multimodal fusion network that combines the features from the two subnetworks to make the final prediction. The model uses a graph representation of the hospital admissions, where each node corresponds to one admission and the edges represent the similarity between admissions measured by the Euclidean distance between patient features. 
Limitations of MM-STGNN include: (i) constrained to each image as one time step and (ii) ignoring the non-linearity of the time between analysis.

Miao \textit{et al.} \cite{Miao2023} proposed the Multimodal Spatiotemporal Graph-Transformer (MuST), that integrates the three modalities: EHR, medical images, and clinical notes, to predict hospital readmissions. The model uses Graph Convolution Networks and temporal transformers to capture spatial and temporal dependencies in EHR and chest radiographs and then a fusion transformer to combine these spatiotemporal features with the features from clinical notes extracted by BioClinical BERT \cite{Alsentzer2019}. The model achieved an AUROC score of 0.858. This model presents significant advancement in hospital readmission prediction by integrating multiple modalities and capturing the complex relationships between patients, admissions, and medical data.


\section{Proposed Approach}
\label{sec:imple}
This section starts by analyzing the selected dataset, followed by the description of the pre-processing steps adopted in order to standardize the data. Finally, we describe the our readmission model, which is based on GNNs.

\subsection{Dataset}
\label{section:dataset}

The dataset used in this work is the MIMIC-IV dataset, version 2.2 \cite{Johnson2023}, which was acquired at the Beth Israel Deaconess Medical Center (BIDMC) in Boston, Massachusetts, USA. The dataset consists of three modules: \textit{hosp}, \textit{icu}, and \textit{note}. The \textit{hosp} module comprises of a decade of admissions between 2008 and 2019, with 431,231 stays and 299,712 patients. From this module we selected the following data: \textbf{Admissions, Diagnoses\_icd, Procedures\_icd, LabEvents, and Patients} tables. The \textit{note} module contains 331,794 deidentified discharge summaries from 145,915 patients. The \textit{icu} module was not used in this work.

\subsection{Data Analysis and Pre-processing}
\label{section:data-analysis}
The \textbf{Admissions} table is the core of the dataset, as it contains information about the patient's admission and his/her journey. Each admission is assigned an id and is associated to an admit and discharge dates. Additional information include the type of admission, the location of the admission and discharge, the insurance type, the language spoken by the patient, the marital status, and the ethnicity of the patient, as well as other demographic features. Since we wanted to combine EHR with clinical discharge notes, we decided to keep only admissions that are associated to a discharge summary. We also filtered the dataset to remove admissions where the patient died at the hospital. In the end, this led to a subset containing 137,769 patients and 303,571 admissions (48.9\% male and 51.1\% female). Additional pre-processing included: i) one hot encoding the categorical features; ii) calculate the length of stay and the time passed since the last admission; and iii) identify readmissions within 30 days. This led to a vector of dimension 78, comprising the features summarized in Table \ref{tab:tab_data}

\begin{table}[t]
	\centering
	\caption{Features used in the model.}
	\resizebox{\textwidth}{!}{
    \begin{tabular}{|c|ccc|}
		\hline
  		Feature  &  Type & Unique Values & Mean (std) \\ \hline
		Age & Numerical & 73 & 59.10 (18.13) \\ 
		Gender & Categorical & 2 & 52/48 (F/M)\\
		Month of Admission & Numerical & 12 &  \\
		Admission Type  & Categorical & 10 &  \\
		Admission Location & Categorical & 12 & \\
		Discharge Location& Categorical & 15 &  \\
		Insurance & Categorical & 3  & \\
		Language & Categorical & 2  & \\
		Ethnicity & Categorical & 6  & \\
		Marital Status & Categorical & 6  & \\
		Length of Stay (Hours) & Numerical & & 126.55 (160.88)  \\
		Previous Admission (Days)& Numerical & & 174.59 (437.43) \\
		Previous Admission Type& Categorical & 10& \\
        Clinical Procedures & Categorical & 12175& \\
		Diagnoses & Categorical & 25169& \\
		Lab Events & Categorical & 856& \\
		Clinical Notes & Free-Text& & \\ \hline
	\end{tabular}
    }
\label{tab:tab_data}
\end{table}

The \textbf{Diagnoses} table contains ICD-9 and ICD-10 codes describing the patient's conditions during the stay, while the \textbf{Procedures} table contains ICD-9 and ICD-10 codes for the procedures the patient underwent. The Diagnoses table contains a total of 3,788,725 diagnoses for the cohort described above, with an average number of diagnoses per admission equal to 12.49 $\pm$ 7.14. For the Procedures table, there are 509,275 procedures, corresponding to 175,424 admissions from 96,885 patients. The average number of procedures per admission is 2.90 $\pm$ 2.68. Since both tables are sparse, we decided to select the first 10 diagnoses and 5 procedures for each admission and convert the ICD codes to text. Then, the BioClinicalBERT model \cite{Lee2019} was used to extract embeddings for each of the selected codes. Finally, the embeddings were averaged to create a single vector that represents the admission's diagnoses and procedures. The final dataframes contain 303,439 admissions and 768 features for the Diagnoses table, and 133,343 admissions and 768 features for the Procedures table. For the admissions that lacked one or both features, we replaced them by vectors of zeros.


The \textbf{Lab Events} table contains the results of all laboratory measurements made for each patient, including hematology measurements, blood gases, chemistry panels, and less common tests such as genetic assays. This table comprises 49,179,082 lab events, from 130,632 patients and 286,907 admissions. There are 856 different items in the table being used. The average number of lab events per admission is 171.41. The lab events contain temporal data, as the same item can be measured multiple times during the patient's stay, however, in this work, due to the quantity of lab events and the available resources, it was decided to calculate the percentage of abnormal values for each item in each admission. This led to a final dataframe containing 286,907 admissions and 856 features. In this case we assumed that admissions missing lab events meant that these were supposed to be whithin the normal values, thus we have replaced the missing cases by a vector of 0s.

The last type of used data are the  \textbf{Clinical Notes}. These notes correspond to the discharge summaries of the patients, that as mentioned above, are free-text long form narratives about the patient's stay. There are 331,793 discharge notes, for 145,914 patients. The average length of the discharge notes is 10,550 $\pm7.70$ characters, which is equivalent to about 4 pages of text. The pre-processing of the discharge notes used the BioClinicalBERT model to extract embeddings, as the diagnoses and procedures. Due to the length of the notes, it was used a sliding window technique to handle the long notes using 512 tokens and a  stride of 256 tokens. The resulting embeddings were averaged to create a single vector that represents the admission's discharge notes.

\begin{figure}[t]
    \centering
    \includegraphics[width=0.99\linewidth]{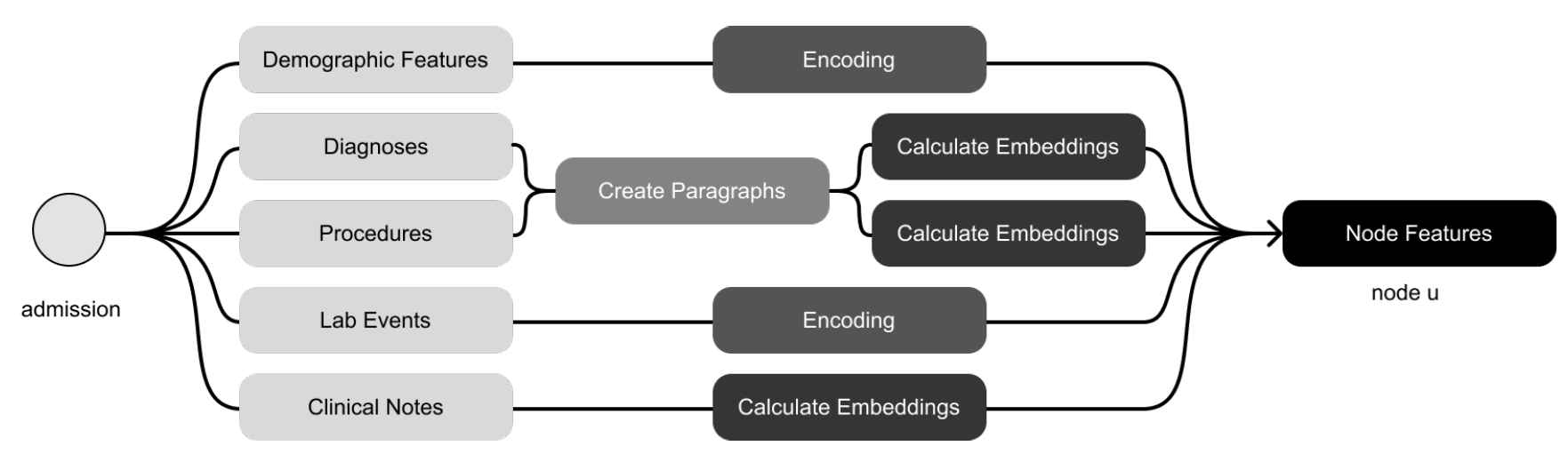}
    \caption{Data analysis and pre-processing overview. Demographic features are encoded into a fixed length vector. ICD codes from the Diagnoses and Procedures tables were converted to text and sent to BioClinicalBERT to obtain the embeddings. Lab events were encoded into a fixed length vector. Clinical notes were sent to BioClinicalBERT to extract embeddings. The node features collect all the vectors from each group.}
    \label{fig:overview}
\end{figure}

We combined all data into a single dataset, containing 303,571 admissions and 3,230 features. The overall pre-processing pipeline is summarized in Fig. \ref{fig:overview}.

\subsection{Readmission Model}
Our readmission model independently analyzes each hospital admission and predicts whether the patient will return in a 30-day period. Contrary to other works \cite{Jamei2017,Shang2021,Alvarez-Romero2022,Garg2021}, we are not limiting our methodology to a cohort of patients that share similar disease(s). This results into a large volume of patients (over 100k) with significant heterogeneity between them. To overcome this issue, we will adopt a strategy based on GNNs, which have shown great promise in this field.

Each admission is considered a node in the graph, $v$, with the features being the demographic data, the diagnoses and procedures embeddings, the lab events, and the discharge notes embeddings. The edges $e$ between admissions translate similarities between them, meaning that we assume two nodes $v_i$ and $v_j$ to be neighbors if the distance between their features falls within a given threshold. 

Due to the large number of nodes (303,571), we used the FAISS library \cite{douze2024faiss} to calculate the similarity between them. FAISS uses a range search to find the neighbors that are within a certain distance of a threshold, which is more efficient than calculating the similarity between all the nodes. 

The choice for the threshold value was critical to balance the trade-off between computational efficiency and the quantity and quality of the edges. To ensure that the model would capture the local structure of the data, without introducing excessive noise, the threshold was tested with 0.8, 0.9, 0.95, and 0.99. The similarity was calculated using the cosine similarity, as it is a common metric used in the literature to calculate the similarity between embeddings.

With all the features, and including self-loops, the edges and degree of the graph obtained were the following:
\begin{itemize}[noitemsep]
    \item Threshold 0.8: 9,256,405 edges, 30.49 average degree;
    \item Threshold 0.9: 3,416,802 edges, 11.26 degree;
    \item Threshold 0.95: 1,237,059 edges, 4.0750 average degree;
    \item Threshold 0.99: 340,539 edges, 1.1218 average degree;
\end{itemize}

The threshold of 0.9 was chosen as the best value, as it provided a good balance between the number of edges and the number of nodes, ensuring that nodes were connected to a sufficient number of similar nodes. 

The architecture used in this work is the Inductive Representation Learning on Large Graphs (GraphSAGE) \cite{Hamilton2018}, since it allows for large scale graph learning and is able to aggregate information from neighbor nodes in a scalable way, which considering the size of the dataset and the available resources, is essential. The GraphSAGE has two components: the aggregator and the update function. The aggregator gathers information from the local neighbors of each node, while the update function updates the node's representation based on the aggregated information. In this work, we compared the following aggregators: mean, max, and add. Our end goal is to perform node prediction, by classifying each node in the graph was resulting in a hospital readmission after 30 days or not. Thus, our model contains a final layer with a sigmoid activation function was used to predict 30-day readmision after the graph processing stage. 


\section{Experimental Setup}
\label{section:experimental_setup}
In this section we start by describing the adopted experimental framework, followed by the description of the training and evalution protocol.

\subsection{Experimental Framework}
 Our readmission prediction model integrates a variety of information regarding each hospital admission. Thus, we have conducted ablation studies in order to evaluate the importance of each group of features in the model's. The admission features were always used, as they contain the demographic data as well as other relevant information about the type of admission. These features were then combined with the remaing, as follows:
\begin{itemize}[noitemsep]
    \item Admission features, Diagnoses and Procedures embeddings, Lab Events;
    \item Admission features, Diagnoses embeddings, Lab Events, Notes embeddings;
    \item Admission features, Procedures embeddings, Lab Events, Notes embeddings;
    \item Admission features, Lab Events, Notes embeddings;
    \item Admission features, Lab Events;
\end{itemize}

Due to the differences of the features types, we performed a specific scaling and normalization for each type of data individually. We used min-max scalers for the admissions and lab events, and the L2 norm for the embeddings.

We also conduct a comparison with two baseline models, commonly applied in predictive tasks that use EHRs: logistic regression (LR) and the multlayer perceptron (MLP).

\subsection{Model Training and Evaluation}
To train and evaluate the different configurations of our readmission model, we randomly split the dataset of 303,571 admissions into train/val/test splits of 60/20/20$\%$. These splits were done based on the patient id (137,769 unique ids) to avoid data leakage. Additionally, given the high imbalance in the number of samples (only 51,985 readmissions in 30-days), this split was stratified to ensure the same distributions across all sets. An additional 20-fold cross validation was performed after identifying the best model configuration to conduct statistical significance studies.

All GNN models were trained using the PyTorch Geometric library. We selected Adam as the optimizer and the binary cross-entropy for loss function. The models were trained for 150 epochs with an early stopping mechanism to prevent overfitting and accelerate the training process, considering a patience of 10 epochs, based on the validation loss. To balance the classes, class weights were used, as an inversely proportional weight to the class frequency.

In order to find the best hyperparameters for the model, a grid-search was performed:
\begin{itemize}[noitemsep]
	\item Learning rate: 0.001, 0.0001, 0.00001,0.000001;
	\item Number of layers: 2, 3, 4;
	\item Hidden units: 32, 64, 128;
    \item Aggregators: max, mean, and add.
\end{itemize}

The baseline models were trained and evaluated using the same data splits, weight classes and cross-validation as the GraphSAGE model. 
All models were evaluated using the following metrics: balanced accuracy (BAcc) and the area under the receiver operating curve (AUROC).

\section{Results and Discusssion}
\label{chapter:results}
First we will focus on the performance of GraphSage and then we will analyze the impact of the different types of data that can be used to characterized a node in the graph. 

\textbf{GraphSAGE Analysis:} We tested multiple hyperparameter configurations for GraphSAGE, as introduced in Section \ref{section:experimental_setup}. Our results showed that changing the hyperparameters did not have a significant impact on the performance, with most of the configurations achieving an AUROC between 0.68 and 0.72 and a BAcc between 0.63 and 0.67. THe best results were obtained with the hidden dimension of 64, 2 layers, learning rate of 1e-05, and mean aggregator. This led to an AUROC of 0.727 and a BAcc of 0.667.

\textbf{Ablation studies:} After finding the best hyperparameters for GraphSage, we performed ablation studies to understand the importance of each type of data for the prediction. The admissions table is the basis of the dataset, so it was included in all the experiments.
The other tables were tested and combined between them to assess how each combination affects the prediction. The experiments performed and the results are presented in Table \ref{table:feature_importance_results}.

\begin{table*}[t]
	\centering
	\caption{Performance comparison using different data combinations. The results are sorted by AUROC scores.}
  \label{table:feature_importance_results}
	\begin{tabular}{|ccc|}
		\hline
		Data & AUROC & Bal. Acc \\ \hline
        All & 0.727 & 0.667\\ 
        Admissions, Procedures, Diagnoses, clinical notes  & 0.719 &  0.660 \\
        Admissions, Diagnoses, Lab Events, clinical notes & 0.707 &  0.650 \\
		Admissions, Procedures, Lab Events, clinical notes & 0.705 & 0.650 \\
		Admissions, Procedures, Diagnoses, Lab Events  & 0.704 &  0.649 \\
		Admissions, Lab Events, clinical notes  & 0.699 &0.645 \\
		Admissions, Lab Events & 0.691 & 0.639 \\ \hline
	\end{tabular}
\end{table*}

These results show that lab events seems to be the least informative table for predicting hospital readmissions, as removing this data has a smaller impact in the BAcc and when used alone or with the clinical notes leads to worse performances.  This may be related to the pre-processing applied to the lab events table, that was encoded to represent the percentage of abnormal results. This consisted of a sparse vector with 856 columns, which may have affected the performance of the model. Procedures, Diagnoses, and Clinical Notes all seem to have a relevant contribution. Overall, the integration of all information led to the best results, suggesting that all of them have some predictive value. 

\textbf{Baseline comparison:} In order to validate the results of the best model, a 20-fold cross-validation was performed to be able to compare the results with the baseline models. The results of the cross-validation are presented in Table \ref{table:baseline_results}. The proposed model significantly outperforms LR, while also achieving better scores than the MLP.

\begin{table}[t]
  \centering
  \caption{Comparison of the best model with the baseline models using 20-fold cross-validation.}
  \label{table:baseline_results}
  \begin{tabular}{c|cc}
    Model & AUROC. &  BAcc \\
    \hline
    GraphSAGE & 0.7163 &  0.6585 \\
    MLP & 0.7089 &  0.6501 \\
    LR & 0.6729 & 0.6298 
  \end{tabular}
\end{table}

After performing the 20-fold cross-validation, we compared the performance of the three models and assessed the statistical significance of the results . We started by testing if the distribution of the metrics was normal, using the Shapiro-Wilk test. The pvalue results are presented in Table \ref{table:shapiro_wilk_test}. All tests were above 0.05, indicating that the distributions are normal. Then, each metric was compared between the models using the T-test. The results are presented in Table \ref{table:t_test_results}. To be considered statistically significant, the p-value must be below 0.05. The results show that the GraphSAGE model has a significantly better performance than both baselines. Additionally, the MLP model has a significantly better performance than the Logistic Regression model for all metrics.

\begin{table}[t]
	\centering
	\caption{Shapiro-Wilk test of the GraphSAGE, Logistic Regression, and Multi Layer Perceptron models.}
  \label{table:shapiro_wilk_test}
	\begin{tabular}{c|cc}
   Model & AUROC & BAcc. \\
    \hline
    GraphSAGE & 0.25358  &  0.9189 \\
    LR & 0.63821  &  0.91510 \\
    MLP & 0.14059  &  0.15461 \\
	\end{tabular}
\end{table}

\begin{table*}[t]
  \centering
  \caption{T-test results for the AUROC, AUPRC, and Balanced Accuracy of the GraphSAGE, Logistic Regression, and Multi Layer Perceptron models.}
  \label{table:t_test_results}
  \begin{tabular}{|c|cc|cc|}
    \hline
    \multirow{2}{*}{Models} & \multicolumn{2}{c|}{AUROC} & \multicolumn{2}{c|}{Bal. Acc.} \\ 
    \cline{2-5}
           & statistic & pvalue &  statistic & pvalue \\ \hline
    GraphSAGE vs. LR & 17.3218 & 4.2701e-13 &  14.3049 & 1.2643e-11 \\
    GraphSAGE vs. MLP & 4.8672 & 0.0001 & 7.0111 & 1.1222e-06 \\
    LR vs. MLP & -19.8683 & 3.5912e-14 & -11.3001 & 7.0992e-10 \\
    \hline
  \end{tabular}  
\end{table*}

\textbf{Comparison with SOTA works:} Compared to the works in the literature that also used GNNs, our results are in the same range of others. MM-STGNN \cite{Tang2023}, reported an AUROC of 0.79, and MuST \cite{Miao2023} reported an AUROC of 0.85, both also using the MIMIC-IV dataset and GNNs. However, these models take advantage of medical images and explore the temporality of the data, which is not the case of this work. Additionally, they work with a significantly less admissions (in the order of 10,000) and consequently having a much smaller graph than the one from this thesis (with approximately 300,000 nodes). 

\textbf{Limitations:} The main limitation of this work pertains the handling of missing data for several of the admissions. Since we filled the missing values with zeroes, it would be interesting to explore different ways to address the missing values - a possibility will be to separate the different features (admissions, procedures, etc) into separate nodes in the graph, instead of combining all the info into a single patient node.
As seen in Table \ref{table:feature_importance_results}, the lab events table was the least important for the prediction of hospital readmissions. This is likely due to the pre-processing used in this work. In a future work, it would be interesting to explore different ways to pre-process this table, and understand if the lab events are actually important for the prediction.

\section{Conclusions}
\label{chapter:conclusions}

This work aimed to predict 30-day hospital readmission using the MIMIC-IV dataset, clinical notes, and graph neural networks. The main goal was to process the data and implement a GNN model to predict hospital readmissions. The admission information includes: patients' demographics, diagnoses and procedures codes, lab results, and discharge notes contain the clinical notes of the patients.

Each node of the graph collects the admission information, while the edges represent the similarity between admissions. Our network architecture is a two-layer GraphSAGE where the similarity was calculated using cosine similarity with the FAISS library. Among the evaluated aggregators (max, mean, and add), the mean aggregator provided the best results, and the lab results correspond to the less informative data. The best model was able to achieve an AUROC of 0.7269 and a balanced accuracy of 0.6668. These results are promising and demonstrated the potential of using GNNs to understand the complex relationships between the features of the MIMIC-IV dataset and predict hospital readmissions.

Some suggestions for future work consist of exploring the temporality of the data by using subgraphs for each patient and their admissions; using a stronger model, such as GCN or GAT, to improve the results; creating a different graph structure, using each type of feature as a different node; using a different approach to fill the missing values, such as using a model to predict the missing values; using the diagnoses and procedures codes in a different way, instead of using them as embeddings; and pre-processing the lab events data in a different way, to extract more information from the lab results.

%
\bibliographystyle{splncs04}

\bibliography{refs.bib}

\end{document}